\documentclass{article}
\usepackage{ijcai21}

\usepackage{times}
\usepackage{soul}
\usepackage{url}
\usepackage[hidelinks]{hyperref}
\usepackage[utf8]{inputenc}
\usepackage[small]{caption}
\usepackage{graphicx}
\usepackage{amsmath}
\usepackage{amsthm}
\usepackage{booktabs}
\usepackage{algorithm}
\usepackage{multirow}
\usepackage{multicol,makecell}

\usepackage{url,graphicx,tabularx,array, booktabs, mathrsfs, color, epstopdf,tabu, svg, booktabs} 
\usepackage{amsmath, bm, amsfonts, cases, algpseudocode, enumitem}
\usepackage[switch]{lineno}
\usepackage{dsfont,natbib}

\title{Representation Memorization for Fast Learning New Knowledge without Forgetting}

 \author{
Fei Mi$^1$\footnote{This work was mainly done when Fei Mi was a Ph.D. in EPFL},
Tao Lin$^2$, \and
Boi Faltings$^2$ \\
$^1$Huawei Noah's Ark Lab, \\
$^2$Swiss Federal Institute of Technology in Lausanne (EPFL)\\
\texttt{mifei2@huawei.com, \{tao.lin,boi.faltings\}@@epfl.ch}}

\newcommand{\MbPA}{\emph{MbPA}}
\newcommand{\Mixture}{\emph{Mixture}}
\newcommand{\Parametric}{\emph{Parametric}}
\newcommand{\Hebb}{\emph{Hebb}}

\begin{document}

\maketitle

\begin{abstract}
The ability to quickly learn new knowledge (e.g. new classes or data distributions) is a big step towards human-level intelligence.
In this paper, we consider scenarios that require learning new classes or data distributions quickly and incrementally over time, as it often occurs in real-world dynamic environments.
We propose ``Memory-based Hebbian Parameter Adaptation'' (\Hebb) to tackle the two major challenges (i.e., catastrophic forgetting and sample efficiency) towards this goal in a unified framework.
To mitigate catastrophic forgetting, \Hebb\ augments a regular neural classifier with a continuously updated memory module to store representations of previous data. 
To improve sample efficiency, we propose a parameter adaptation method based on the well-known Hebbian theory~\cite{hebb2005organization},
which directly ``wires'' the output network's parameters with similar representations retrieved from the memory.
We empirically verify the superior performance of \Hebb\
through extensive experiments on a wide range of learning tasks (image classification, language model) and learning scenarios (continual, incremental, online).
We demonstrate that \Hebb~effectively mitigates catastrophic forgetting, and it indeed learns new knowledge better and faster than the current state-of-the-art.

\end{abstract}

\section{Introduction}
      
      In real-life machine learning applications, new knowledge (e.g., new classes or data distributions) arrive gradually over time. The ability to quickly learn and accumulate new knowledge without forgetting old ones is a hallmark of artificial intelligence. 
    Two major challenges prevent standard neural networks to be applied towards this goal. (1) \textbf{catastrophic forgetting}: continually incorporating new knowledge requires additional training and often reduces the performance on old ones learned before \cite{mccloskey1989catastrophic}; (2) \textbf{sample efficiency}: the amount of data on new knowledge is often limited, which prevents neural networks from being trained to achieve reasonable accuracy \cite{wang2019few}.
    
    The \textit{catastrophic forgetting} challenge is recently studied in the context of  ``continually/incrementally'' learning a sequence of tasks.  Various regularization-base methods \cite{kirkpatrick2017overcoming,rebuffi2017icarl,castro2018end} and memory-based approaches \cite{grave2016improving,merity2016pointer,rebuffi2017icarl} have been proposed to mitigate catastrophic forgetting.
    The \textit{sample efficiency} challenge is recently studied in the context of few-shot learning; a popular approach is ``meta-learning''~\cite{finn2017model} that learns over a bunch of specifically structured meta-tasks. 
    
     However, existing methods often tackle these two challenges separately.  
    To this end, we propose a method called Memory-based Hebbian Parameter Adaptation (\Hebb) to tackle them in a \textit{unified} framework. 
     \Hebb\ makes use of a memory component similar to \cite{merity2016pointer,sprechmann2018memory} that stores previous input representations to mitigate catastrophic forgetting.  
      To improve sample efficiency on new knowledge, we propose a parameter adaptation procedure based on the well-known Hebbian theory \cite{hebb2005organization} during inference. It directly \textit{wires} similar representations retrieved from the memory to the corresponding parameters of the classifier's output network. 
      The memory accessing operation and the parameter adaptation procedure can be easily computed such that \Hebb\ can be easily plugged into different neural classifiers and learning scenarios.
      
Besides the standard continual learning setting \cite{rebuffi2017icarl} to evaluate the ability to mitigate catastrophic forgetting, two other learning scenarios are considered to evaluate the ability to deal with the sample efficiency challenge.
In the first \textit{incremental learning} scenario, a pre-trained classifier is initially trained on a small dataset containing new knowledge. Then it is fixed to be evaluated w.r.t. future observations. 
    In the second \textit{online adaptation} scenario, we have no data on new knowledge initially, and the pre-trained classifier needs to continuously learn new knowledge through a single pass over new data in an online manner. 
    These two learning scenarios are both practically critical. For example, the incremental learning scenario could simulate that a robot is shown some images of new objects before they appear in its routine tasks.
 In the online adaptation scenarios, a robot has to deal with new objects continuously.

    Through extensive experiments on a wide range of learning tasks (image classification, language model) and learning scenarios (continual, incremental, online), we empirically demonstrate: (i) \Hebb\ can be easily plugged into different neural classifiers and learning scenarios with trivial computation overhead. (ii) \Hebb\ effectively mitigates catastrophic forgetting, indicated by its superior performance compared to \MbPA~\cite{sprechmann2018memory} and \emph{EWC} \cite{kirkpatrick2017overcoming} in the continual learning setting. (iii) More importantly, \Hebb\ notably improves sample efficiency for fast learning new classes and data distributions. It outperforms various state-of-the-art methods in both incremental learning and online adaptation scenarios, especially on new or infrequent classes.

      \section{Related Work}
      \subsection{Memory-augmented Neural Networks}
      Recently, various memory modules ($\mathbf{M}$) have been proposed to augment neural networks for remembering long-term information~\cite{graves2014neural,grave2016improving,grave2017unbounded,merity2016pointer} or learning infrequent patterns \cite{santoro2016one,kaiser2017learning,sprechmann2018memory,mi2020mem}.
      
      There are many variants of how to read from $\mathbf{M}$ and mix the entries retrieved from $\mathbf{M}$ with the network computation.
      One approach is through some differentiable context-based lookup mechanisms~\cite{graves2014neural,santoro2016one} for learning to match the current activation to past activations stored in $\mathbf{M}$.  
      However, these mechanisms often require strong memory supervision, and the size of the $\mathbf{M}$ has to be fixed.
      Another approach is using a simple mixture model. In this case, a non-parametric prediction is computed based on the similarity between the entries in $\mathbf{M}$ and the current input.
      The neural network's prediction is directly interpolated with the non-parametric prediction from $\mathbf{M}$. 
      This approach has been shown simple but effective for language modeling~\cite{grave2016improving,grave2017unbounded}, neural machine translation~\cite{tu2018learning}, image classification~\cite{orhan2018simple}, and recommendation~\cite{mi2020mem}.
      Recently,~\cite{sprechmann2018memory} introduces \MbPA\ to use nearest neighbors retrieved from $\mathbf{M}$ for parameter adaptation during model inference for the fast acquisition of new knowledge. 
      \MbPA++ \cite{d2019episodic} improves \MbPA\ to better mitigate catastrophic forgetting through better memory management during training. 
      The framework proposed in this paper is motivated by~\cite{sprechmann2018memory}, and it mainly improves \MbPA\ for better learning new knowledge. 
      
      \subsection{Hebbian Learning}
      
     Hebbian theory~\cite{hebb2005organization} is a neuroscientific theory attempting to explain ``synaptic plasticity'', i.e., the adaptation of brain neurons during the learning process. For artificial neural networks, Hebbian theory describes a method of determining how to alter the weights between two neurons. 
         It is also related to early ideas from psychology and neuroscience, called \emph{associative memory}.
      In psychology, associative memory is the ability to learn and remember the relationship between unrelated items. 
      In neuroscience, associative memory means that the information is stored by associative structures to bind representative patterns to their corresponding concepts or labels. 
      
          Recent approaches apply Hebbian theory to every single neuron connection for fast network weight learning.
      \citet{ba2016using} proposes a \textit{fast weight} to augment the standard computation of RNNs. The fast weight is defined as the running average of the outer product of two hidden states in RNNs. It is multiplied to the current state and it is continuously updated to allow each new hidden state to be attracted to recent hidden states. 
      \citet{miconi2018differentiable} later augments the traditional connection between two neurons in general neural networks with a \textit{Hebbian trace}. The Hebbian trace between two neurons is defined as a running average of the scalar product of the first neuron's activation in the last timestamp and the second neuron's activation in the current timestamp.
   The Hebbian trace is merged with the standard neuron connection through a \textit{differentiable plasticity} optimized by SGD.
      The later extension~\cite{miconi2018backpropamine} introduces a term parametrized by another neural network to learn how fast should new information be incorporated.
      
 Instead of applying the Hebbian theory to every single neuron connection, we use it to directly wire the activation input to the output layer with the corresponding class label for fast binding new classes. 
Similar perspectives are recently proposed. For example, \citet{munkhdalai2018metalearning} augments the layer preceding the Softmax layer with the Hebbian updates followed by a nonlinear activation for meta-learning.
    \citet{rae2018fast} proposes a Hebbian Softmax layer during the normal model training phase to better learn infrequent vocabularies in language modeling tasks.
    The Hebbian update rule proposed in our paper is motivated by \citet{rae2018fast}, yet our Hebbian update rule is only applied to relevant entries retrieved from a continuously updated memory module for the purpose of fast learning new knowledge in an incremental or online manner.

      \section{Memory-based Hebbian Parameter Adaptation}
     
   This section introduces the Memory-based Hebbian Parameter Adaptation (\Hebb) method to help standard neural classifiers mitigate catastrophic forgetting and improve sample efficiency.
    First, we introduce a memory component to store representations of past data. Then, we introduce the Hebbian update for fast learning new knowledge during inference, and we compare it with state-of-the-art (\MbPA~\cite{sprechmann2018memory}). Lastly, we introduce a dynamic interpolation of the proposed Hebbian update and \MbPA.
      
\paragraph{\textbf{Background}} Neural classifiers can be visualized by two parts. The first part is a \textit{feature extractor} $g_{\theta}$ to compute a \textit{input representation vector} $\mathbf{h}_x \!=\! g_{\theta}(x)$ for an input $x$. The second part is an \textit{output network} $f_\omega$ for predicting $\hat{y} \!=\! f_\omega(\mathbf{h}_x)$. A fully-connected layer with a Softmax activation is often used:  $f_\omega(\mathbf{h}_x) \!=\! \text{Softmax} (\mathbf{W}^\top \mathbf{h}_x + \mathbf{b})$. The weights $\mathbf{W} \in \mathbb{R}^{d \times n}$ and the bias $\mathbf{b} \in \mathbb{R}^n$, where $d$ is the dimension of $\mathbf{h}_x$ and $n$ is the number of classes.

    \subsection{Memory Component}
\label{subsec:memory}
      Motivated by~\cite{grave2016improving,sprechmann2018memory}, we design a memory module $\mathbf{M}$ in the form of key-value pairs, i.e. $\mathbf{M} = \{(key, value)\}$, to tackle the catastrophic forgetting challenge. $\mathbf{M}$ is indexed by keys, and we define keys to be input representations while \textit{values} are the corresponding class labels. Storing input representations rather than raw inputs also helps to preserve data privacy.
      Upon observing a training data $(x, y)$, we write a new entry to $\mathbf{M}$ by:
      \begin{equation}
      \left\{
      \begin{aligned}
       key &  \leftarrow \mathbf{h}_x \!=\! g_\theta (x) \\
      value & \leftarrow y
      \end{aligned}
      \right. \,
      \label{eq:memory}
      \end{equation}

    To scale up to a large number of observations in practical scenarios, we utilize the FAISS library\footnote{\url{https://github.com/facebookresearch/faiss}} to implement a scalable retrieval method with Product Quantization~\cite{jegou2010product} to achieve both computation and storage efficiency. Settings with limited and unrestricted memory sizes are both considered in later experiments.
    
       To adapt the prediction for an input $x$ during inference, we retrieve a set of $K$ \textit{nearest} neighbors of its representation $\mathbf{h}_x= g_{\theta}(x)$ from $\mathbf{M}$ by:
   \begin{equation}
   \mathbf{N} = \{ (\mathbf{h}_k, y_k, c_k) \}_{k=1}^K,
   \end{equation}    
   where $c_k$ is the closeness between $\mathbf{h}_x$ and $\mathbf{h}_k$, and we use the same kernel function $c_k = \frac{1}{\epsilon + ||\mathbf{h}_x - \mathbf{h}_k||^2_2}$ as in~\cite{sprechmann2018memory}. Entries in $\mathbf{N}$ are used to adapt the parameters of $f_\omega$ and details are introduced next.

\subsection{Hebbian Update}
    
       The general Hebbian theory \cite{hebb2005organization} is expressed as:
       \begin{equation}
           W[i,j] = \frac{1}{n} \sum_{k=1}^n x_i^k x_j^k,
       \end{equation}
       where $W[i,j]$ is the weight of the connection from neuron $i$ to neuron $j$; $x_i^k$ is the k-th input to the neuron $i$, and similarly for $x_j^k$; $k \in {1...n}$ and $n$ is the number of training samples. Therefore, the multiplication of $x_i^k$ and $x_j^k$ summing over $n$ training examples gives the weight $W[i,j]$ between the neuron $i$ and $j$. The intuition is: if nodes $i$ and $j$ are often activated together, they have a strong connection weight. 
       
      Next, we propose a Hebbian update rule using the above Hebbian theory to adapt the output network $f_\omega$ for fast-learning new classes.
      The weight $\mathbf{W} \in \mathbb{R}^{d \times n}$ of $f_\omega$ can be seen as a set of $n$ vectors $\mathbf{w}_i \in \mathbb{R}^d$, where $i \in \{1, \ldots, n\}$ with each $i$ corresponds to a class.
      The Hebbian update rule for $\mathbf{w}_i$ is defined as:
      \begin{equation}
      \Delta^{Hebb}_{\mathbf{w}_i}  = \frac{1}{|\mathbf{N}_i|} \sum_{k=1}^{|\mathbf{N}_i|}c_k \mathbf{h}_k \,,
      \label{eq:hebbrule}
      \end{equation}
      where $\mathbf{N}_i$ is the subset of entries in $\mathbf{N}$ with class label $i$, $c_k$ is used for weighted update, and $\frac{1}{|\mathbf{N}_i|}$ in Eq. \eqref{eq:hebbrule} averages the cumulative effect of multiple entries with the same class label.
       A similar Hebbian update for  the $i$-th element ($b_i$) of the bias term of $f_\omega$ is: $\Delta^{Hebb}_{b_i}  = \frac{1}{|\mathbf{N}_i|} \sum_{k=1}^{|\mathbf{N}_i|}c_k $.
               
          The Hebbian update rule in Eq. (\ref{eq:hebbrule}) applies the principle of Hebbian theory to the output layer by directly \textit{wiring} the input representation $\mathbf{h}_k$ and the corresponding label $y_k$ together, where $W$, $x_i^k$, $x_i^k$ in the Hebbian theory correspond to our $\mathbf{W}$, $h_k$, $y_k$ respectively.
          The idea is to ``\textbf{memorize}'' representations of a new class in a sample-efficient manner by directly assigning them to the output network's corresponding parameters. With the Hebbian update, the weights corresponding to a new class aligns with its representations to help the model predict the new class.        
      The vectors in $\mathbf{W}$ of which the corresponding classes are not in $\mathbf{N}$ are not affected by the Hebbian update. Therefore, it only does a \textit{sparse update} for parameters relevant to neighbors in $\mathbf{N}$.
      A detailed analysis of the advantage of the proposed Hebbian update is included below.

      \subsection{Analysis and Comparison to Existing \MbPA}
      
\label{subsec:compare}

      The state-of-the-art parameter adaptation method \MbPA~\cite{sprechmann2018memory} is to adapt $f_\omega$ by maximizing the weighted log likelihood w.r.t. $\mathbf{N}$ by:
      \begin{equation}
      \max_{\omega} \mathcal{L}^{\mathbf{N}}(f_\omega) = \max_{\omega} \frac{1}{ |\mathbf{N}| } \sum_{k=1}^{ |\mathbf{N}| } c_k \log P(y_k|\mathbf{h}_k, \omega) \,,
      \label{eq:mbpa1}
      \end{equation}
      where $P_{y_k} \!:=\! P(y_k|\mathbf{h}_k, \omega) $ is the predicted probability on true label $y_k$ for the $k$-th neighbor.
      The objective function is optimized by gradient descent, and one optimization step without considering learning rate can be written as:
      $
      \Delta \omega = - \nabla_{\omega} \mathcal{L}^{\mathbf{N}} (f_\omega)
      $.

    
      With the standard $\text{softmax}$ activation with cross-entropy loss,
      the gradient contributed by the $k$-th neighbor with label $y_k$ w.r.t. to $\mathbf{w}_i$ is $ (P_{y_k} \!-\! \delta(i, y_k)) \mathbf{h}_k$, where $\delta()$ is the Kronecker delta. 
      Therefore, the MbPA update with one optimization step for $\mathbf{w}_i$ can be decomposed as:
      \begin{equation}
      \begin{aligned}
      \Delta^{MbPA}_{\mathbf{w}_i}  & =   \Delta^{\mathbf{N}_i}_{\mathbf{w}_i} + \Delta^{\overline{\mathbf{N}}_i}_{\mathbf{w}_i}\\
      & = - \nabla_{\mathbf{w}_i}  \mathcal{L}^{\mathbf{N}_i} (f_\omega)  - \nabla_{\mathbf{w}_i}  \mathcal{L}^{\overline{\mathbf{N}}_i} (f_\omega)  \\
      & = \frac{1}{|\mathbf{N}|} \sum_{k=1}^{|\mathbf{N}_i|} c_k  (1-P_{y_k})\mathbf{h}_k - \frac{1}{|\mathbf{N}|}\sum_{j=1}^{|\overline{\mathbf{N}}_i|} c_j P_{y_j}\mathbf{h}_j \,,
      \end{aligned}
      \label{eq:mbpa}
      \end{equation}
      where we decompose $\mathbf{N}$ into two disjoint sets $\mathbf{N}_i$ and $\overline{\mathbf{N}}_i$.
      $\mathbf{N}_i \!=\! \{ (\mathbf{h}_k, y_k \!=\! i, c_k) \}_{k=1}^{|\mathbf{N}_i|}$ contains entries with label $i$, and $\overline{\mathbf{N}}_i \!=\! \{ (\mathbf{h}_j, y_j \! \neq\! i, c_j) \}_{j=1}^{|\overline{\mathbf{N}}_i|}$ contains entries with label different from $i$. $P_{y_k}$ is the predicted probability on the label $y_k$ of the $k$-th neighbor and similarly for $P_{y_j}$.
    $\Delta^{\mathbf{N}_i}_{\mathbf{w}_i}$ is the update from entries with label $i$, and $\Delta^{\overline{\mathbf{N}}_i}_{\mathbf{w}_i}$ is the update from other entries.
 
 For a new class $i$, effectively updating $\mathbf{W}$ towards representations of $i$ is crucial for fast-learning this class. 
Next, we analyze why $\Delta^{Hebb}_{\mathbf{w}_i} $ can better learn new classes than $ \Delta^{MbPA}_{\mathbf{w}_i} $ through the lens of the two terms in $ \Delta^{MbPA}_{\mathbf{w}_i}$.
 \begin{itemize} 
 \item \textbf{
$\Delta^{Hebb}_{\mathbf{w}_i} $ is more sensitive than the first term $\Delta^{\mathbf{N}_i}_{\mathbf{w}_i}$ of $\Delta^{MbPA}_{\mathbf{w}_i}$ in terms of memorizing new class representations.
 }
 For a new class $i$, $P_{y_k}$ is often very small because the model is not yet confident on this class ;
 thus, the direction of $\Delta^{\mathbf{N}_i}_{\mathbf{w}_i}$ is \textbf{similar} to the direction of $\Delta^{Hebb}_{\mathbf{w}_i}$.
   Moreover, the magnitude $\Delta^{\mathbf{N}_i}_{\mathbf{w}_i}$ of  is \textbf{bounded} by the magnitude of $\Delta^{Hebb}_{\mathbf{w}_i}$, because 
    $
    |\Delta^{\mathbf{N}_i}_{\mathbf{w}_i} | \!=\! |\frac{1}{|\mathbf{N}|} \sum_{k=1}^{|\mathbf{N}_i|} c_k  (1-P_{y_k})\mathbf{h}_k|
    \!<\!  |\frac{1}{|\mathbf{N}|} \sum_{k=1}^{|\mathbf{N}_i|} c_k  \mathbf{h}_k | \!\leq\!  |\frac{1}{|\mathbf{N}_i|} \sum_{k=1}^{|\mathbf{N}_i|} c_k  \mathbf{h}_k | \!=\! |\Delta^{Hebb}_{\mathbf{w}_i}|
    $.
    Therefore, $\Delta^{Hebb}_{\mathbf{w}_i} $ makes a larger update than $\Delta^{\mathbf{N}_i}_{\mathbf{w}_i}$ towards the representations of a new classes.
\item \textbf{The second term $\Delta^{\overline{\mathbf{N}}_i}_{\mathbf{w}_i}$ of $\Delta^{MbPA}_{\mathbf{w}_i}$ prevents effectively learning a new class $i$.}
    For a new class $i$, many entries in $\overline{\mathbf{N}}_i$ correspond to old classes, therefore, their $P_{y_j}$ are often relatively large and their $\mathbf{h}_j$ could be very different from $\mathbf{h}_k$. In other words, $\Delta^{\overline{\mathbf{N}}_i}_{\mathbf{w}_i}$ is very different from the representations of $i$ such that it prevents updating parameters towards the optimal direction for $i$.
 \end{itemize}
   
            \begin{figure}[t!]
    \centering
    \includegraphics[width=0.49\textwidth]{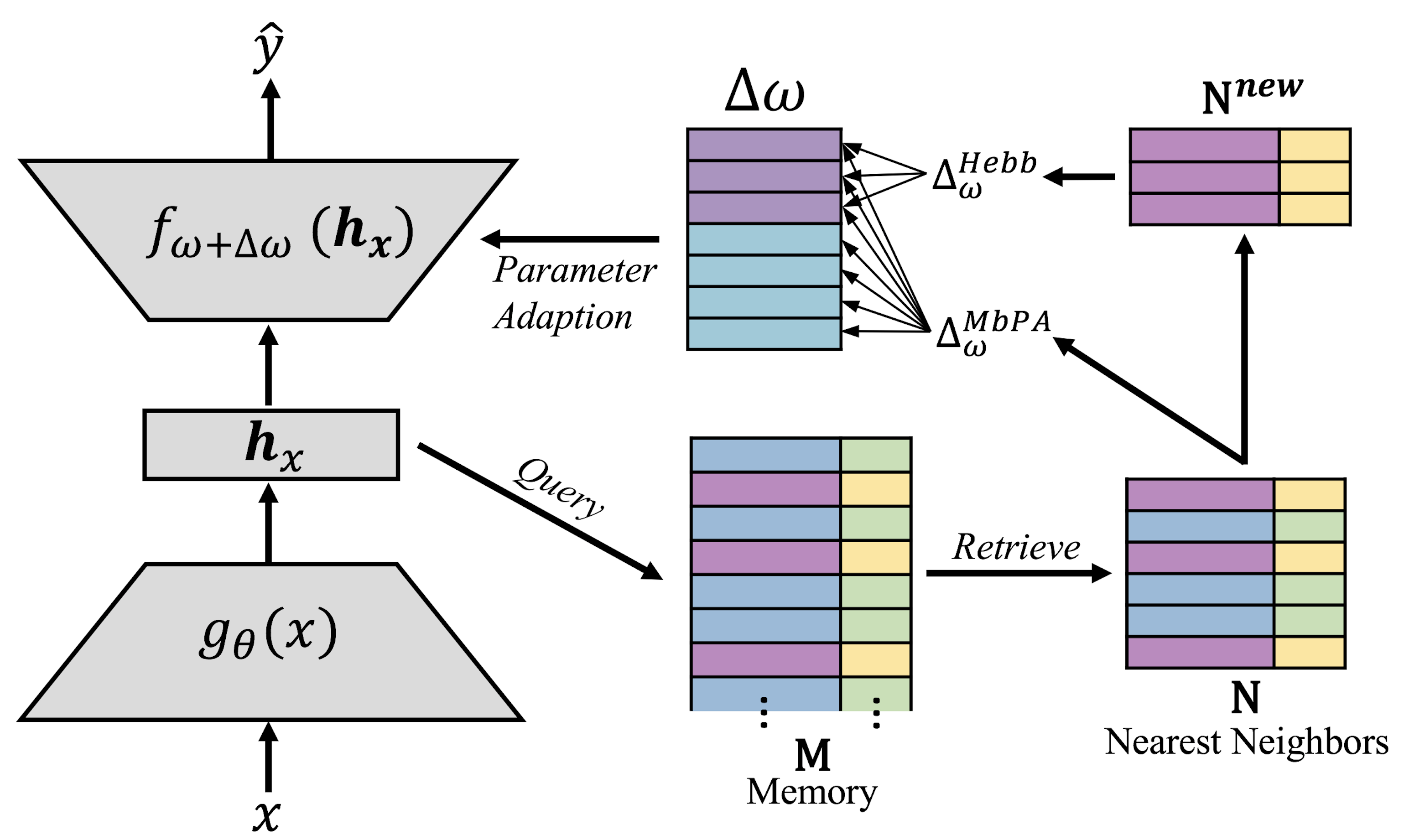}
    \caption{Computation pipeline of \Hebb\ during inference. New classes not in the training phase are colored by yellow and the corresponding representations are colored by purple tone in $\mathbf{M}$. Green values and blue-tone keys in $\mathbf{M}$ correspond to old classes.}
    \vspace{-0.05in}
    \label{fig:model}
\end{figure}

      \subsection{Mixture of Hebbian and \MbPA\ update}
    
     As we analyze in previous subsection that the MbPA update ($\Delta_{\omega_i}^{MbPA}$) in Eq.~\eqref{eq:mbpa} can deal with old classes, while the sparse Hebbian update ($\Delta_{\omega_i}^{Hebb}$) in Eq.~\eqref{eq:hebbrule} is designed to learn new classes fast.
     We propose a mixed update scheme to flexibly work with both scenarios.
     Instead of interpolating them by a fixed weight, 
     we extend the idea of~\cite{cui2019class} for re-weighting loss function for imbalanced classes, and propose a dynamic interpolation by:
      \begin{equation}
      \left\{
      \begin{aligned}
      \Delta\omega_i &= (1-E_i)\Delta_{\omega_i}^{MbPA} + E_i \Delta_{\omega_i}^{Hebb}\\
      E_i & = \frac{1-\beta} {1 - \beta^{n_i} },
      \end{aligned}
      \right. \,.
      \label{eq:interpolation}
      \end{equation}
     where $\omega_i$ is the parameter ($\mathbf{w}_i$ or $b_i$) of $f_\omega$ for class $i \in \{1, \dots, n\}$, and $\Delta\omega_i$ is the hybridized local adaptation for $\omega_i$.
      $n_i$ is the occurrence frequency of class $i$ and $\beta\in [0, 1)$ is a hyper-parameter controlling the decay rate of $E_i$ as $n_i$ increases. 
       The idea is to rely more on the sparse Hebbian update (i.e., $\Delta_{\omega_i}^{Hebb} $) when class $i$ has not been seen many times. As it gradually becomes a frequent class, the adaptation relies more on the MbPA update.

    The final prediction after parameter adaptation (during inference) is computed by
    $
    \hat{y} = f_{\omega + \Delta \omega}(\mathbf{h}_x)
    $. The local adaptation $\Delta \omega$ to $f_\omega$ is discarded after the model makes a prediction, avoiding long term overheads (e.g. overfitting).

    \subsection{\Hebb\ Algorithm}

The training and inference procedures of \Hebb\ are given in Algorithm~\ref{alg:one}, and a detailed computation pipeline during inference is illustrated in Figure~\ref{fig:model}.
Training data representations are stored during training to alleviate catastrophic forgetting, while the fast learning ability is achieved through parameter adaptation during inference.

\begin{algorithm}[t]
    \begin{algorithmic}[1]
     \Procedure{Train}{training data: $D_{train}$}
          \State Train $g_\theta$ and $f_\omega$ w.r.t $D_{train}$ 

          \For {($x, y) \in D_{train}$}
          \State Store $(g_\theta(x), y)$ into $\mathbf{M}$
          \EndFor

        \EndProcedure
        
        \Procedure{Inference}{input: $x$, ground truth: $y$}
          \State Calculate input representation $\mathbf{h}_x = g_\theta(x)$ 
          
          \State Retrieve K-nearest neighbors $\mathbf{N}$ of $\mathbf{h}_x$ from $\mathbf{M}$
          
          \State Compute $\Delta_{\omega}^{MbPA}$ w.r.t. $\mathbf{N}$ 
          
          \State Select $\mathbf{N^{new}}$ and compute $\Delta_{\omega}^{Hebb}$ w.r.t. $\mathbf{N^{new}}$ 
          
          \State  Combine $\Delta_{\omega}^{MbPA}$ and $\Delta_{\omega}^{Hebb}$ by Eq. (\ref{eq:interpolation})
          
          \State Predict $\hat{y} = f_{\omega + \Delta \omega} (\mathbf{h}_x)$ 
          
          \State Store $(\mathbf{h}_x, y)$ into $\mathbf{M}$ if ``online adaptaion''
        \EndProcedure
          \caption{\small Memory-based Hebbian Parameter Adaptation}
          \label{alg:one}
        
    \end{algorithmic}
\end{algorithm}

\begin{itemize}
\item \textbf{Training procedure}: the feature extractor $g_\theta$ and the output network $f_\omega$ are trained first. Then, input representations and their corresponding labels of training data are stored to $\mathbf{M}$.
\item \textbf{Inference procedure}: the set of nearest neighbors $\mathbf{N}$ of the input $x$ is retrieved from $\mathbf{M}$, and $\Delta_{\omega}^{MbPA}$ is computed w.r.t. $\mathbf{N}$.
When there are new classes to be learned after the initial training phase, we select a subset $\mathbf{N^{new}}$ of $\mathbf{N}$ whose labels are not seen during the initial training phase to compute $\Delta_{\omega}^{Hebb}$.
Afterwards, these two updates are combined by Eq. (\ref{eq:interpolation}) before a prediction $\hat{y} = f_{\omega + \Delta \omega} (\mathbf{h}_x)$ is computed.
In the cases of online adaptation during inference, e.g. evaluated in Section \ref{subsec:cifar_online} and \ref{subsec:lm}, $\mathbf{M}$ is continuously updated with new testing data.
\end{itemize}

	     \section{Experiments}

\begin{table*}[htb!]
    \setlength{\tabcolsep}{5pt}
            \centering
            \begin{tabular}{clccccccccc}
                  \toprule
                  &  & \multicolumn{3}{c}{ResnetV1} & \multicolumn{3}{c}{Densenet}  & \multicolumn{3}{c}{MobilenetV2} \\
                  \cmidrule(r){3-5} \cmidrule(r){6-8} \cmidrule(r){9-11} 
                 & Model & Epoch 1 & Epoch 3 & Epoch 10 & Epoch 1 & Epoch 3 & Epoch 10 & Epoch 1 & Epoch 3  & Epoch 10 \\
                  \cmidrule(r){1-1} \cmidrule(r){2-2} \cmidrule(r){3-5} \cmidrule(r){6-8} \cmidrule(r){9-11} 
                  \multirow{ 4}{*}{Overall} & \Parametric & 36.10\%  & 41.38\% & 47.45\% & 34.48\%  & 40.05\% & 45.14\%  & 35.12\%  & 40.64\% & 45.73\%   \\
                  & \Mixture & 38.62\%  & 43.06\%  & 47.74\% & 36.11\%  & 41.74\% & 45.57\%  & 35.72\%  & 41.85\% & 45.95\% \\
                  & \MbPA & 38.04\%  & 45.58\%  & 48.76\% & 36.25\%  & 43.90\%  & 47.53\%  & 36.90\%  & 43.51\% & 48.26\% \\
                  & \Hebb & \textbf{39.04}\%  & \textbf{47.16}\%  & \textbf{49.69}\% & \textbf{37.07}\%  & \textbf{45.80}\%  & \textbf{48.02}\%  & \textbf{37.72}\%  & \textbf{45.85}\% & \textbf{48.69}\%\\
                  
                  \cmidrule(r){1-1} \cmidrule(r){2-2} \cmidrule(r){3-5} \cmidrule(r){6-8} \cmidrule(r){9-11} 
                  
                  \multirow{ 4}{*}{New} & \Parametric & 2.03\%  & 19.15\% & 31.67\% & 1.63\%  & 17.65\% & 29.94\%  & 2.01\%  & 18.05\% & 29.17\% \\
                  & \Mixture & 7.08\%  & 22.83\% & 32.05\% & 6.45\%  & 21.06\% & 30.01\% & 6.95\%  & 22.60\% & 30.85\% \\
                  & \MbPA & 7.01\%  & 27.96\% & 36.89\% & 6.55\%  & 27.01\% & 35.92\%  & 6.85\%  & 27.40\% & 36.39\%  \\
                  & \Hebb & \textbf{10.26}\%  & \textbf{31.92}\% & \textbf{39.01}\% & \textbf{9.75}\%  & \textbf{30.23}\% & \textbf{38.10}\% & \textbf{10.02}\%  & \textbf{31.45}\% & \textbf{38.70}\% \\
                  
                  \bottomrule
            \end{tabular}
            \caption{
                Average Top-1 accuracy of the incremental image classification experiment. For each base neural classifier (ResnetV1, Densenet, MobilenetV2), the Top-1 accuracy of different methods on 50 new classes (\textbf{New}) in testing and on all 100 classes (\textbf{Overall}) are reported at epochs of 1, 3, and 5 respectively.
            } \vspace{-0.05in}
            \label{table:cifar-incre}
      \end{table*}   
As \Hebb\ aims to learn new knowledge fast, the majority of our experiments study this aspect.
We consider three learning scenarios for image classification. The continual learning setting in Section \ref{subsec:mnist} briefly studies the catastrophic forgetting issue, while the incremental learning (Section \ref{subsec:incre}) and online adaptation (Section  \ref{subsec:cifar_online}) experiments study fast learning new classes.
    Section \ref{subsec:lm} studies an online adaptation setting for language model with different types of testing data (intra-domain and cross-domain). 

For fairness, the memory component used by different methods is the same. The number of neighbors while using unlimited memory size is set to 200 for different methods because we found that the performance saturates at this neighbor size.
   Results of all following experiments are averaged over three different random seeds used for data split and model training. Model training details and hyper-parameter settings are included in Appendix \ref{appendix:repo_checklist}. Furthermore, a hyper-parameter sensitivity analysis is included in Appendix \ref{appendix:hyper_sensitivity}.

\subsection{Continual Learning for Image Classification}
\label{subsec:mnist}
In this experiment, we studied a continual learning setting to sequentially learn multiple tasks without forgetting previous ones.
The ``permuted MNIST'' \cite{goodfellow2013maxout} dataset is used.
Each task is given by a different random permutation (i.e., distribution shift) of the pixels of the MNIST dataset. We used a chain
of 20 different tasks (20 different permutations) trained sequentially. Each task contains 10,000 samples.
Different methods are trained 100 epochs for each task and evaluated on all
tasks that have been trained on so far. 
The main challenge of this experiment is to prevent catastrophically forgetting image patterns in previous tasks.

\begin{figure}[!t]
		\centering
		\includegraphics[width=0.41\textwidth]{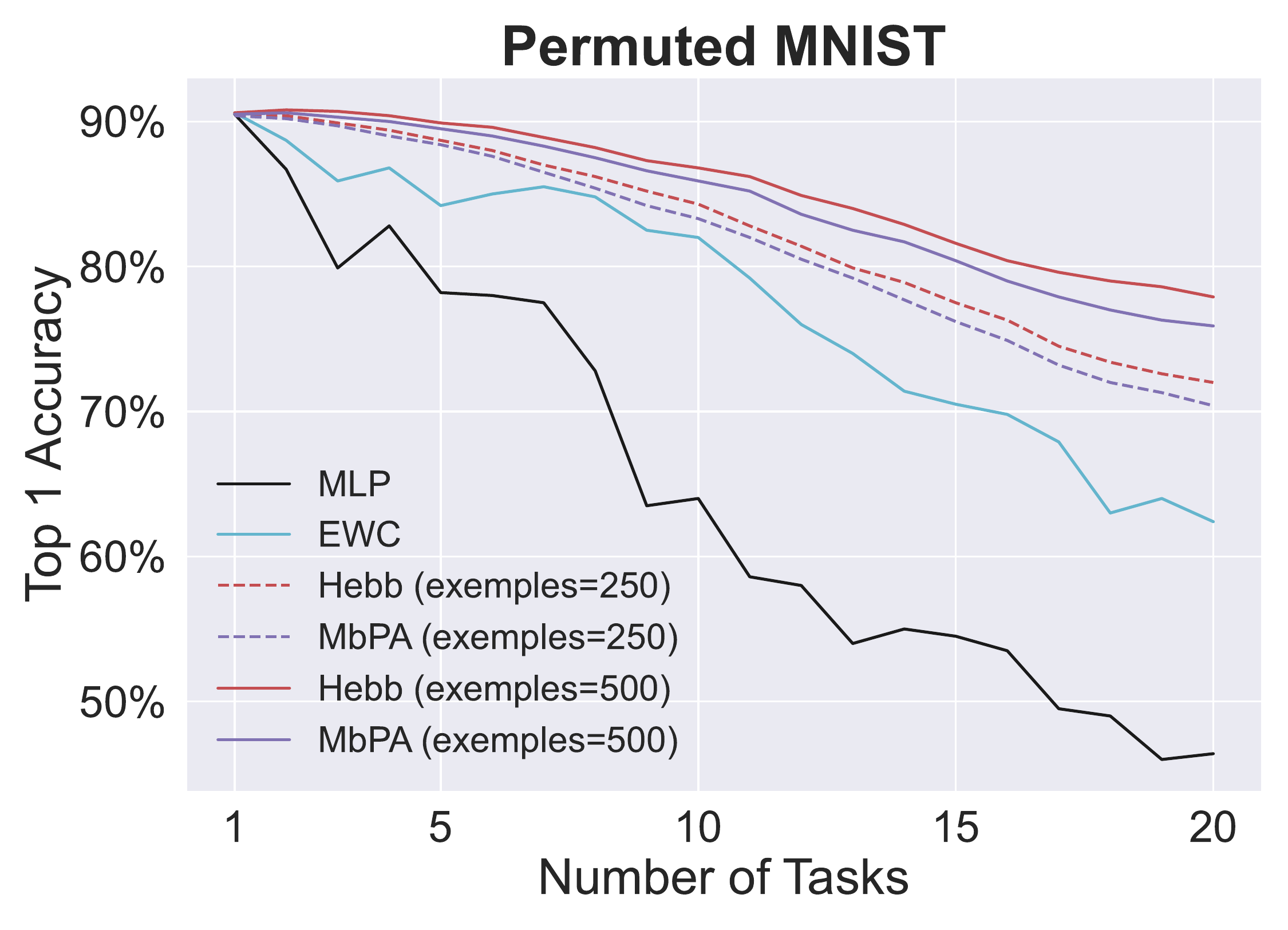} 
       \vspace{-0.1in} \caption{ 
        Continual learning results to learn 20 tasks on the permuted MNIST dataset. 250/500 random samples are stored per task for \MbPA\ and \Hebb.
       \vspace{-0.1in}
        \label{fig:mnist}
        }
\end{figure}

For the base neural classifier, we use a one-layer \textit{MLP} with size 1000.
As in \cite{sprechmann2018memory}, we directly use pixels as input representation to query memory, i.e. an identity function for the feature extractor $g_\theta$, and the \textit{MLP} serves as the output network $f_\omega$.
The Hebbian update in \Hebb\ is applied to all neighbors as no new classes are encountered.
As baseline methods, we compared \Hebb\ with the regular gradient descent training of \textit{MLP}, \textit{EWC} \cite{kirkpatrick2017overcoming}, and \MbPA~\cite{sprechmann2018memory}.

Results comparing different methods are included in Figure~\ref{fig:mnist}.
As an effective approach to alleviate catastrophic forgetting, \textit{EWC} performs much better than \textit{MLP}. Both \Hebb\ and \MbPA\ perform better than \textit{EWC}, which means that local parameter adaptation methods can recover classification performance when a task is catastrophically forgotten.
The better performance of \Hebb\ over \MbPA\ demonstrates that \Hebb\ effectively mitigate catastrophic forgetting.

      \subsection{Incremental Learning for Image Classification}
\label{subsec:incre}
 This experiment studied an incremental learning scenario to learn new knowledge. 
 We considered the image classification task on the CIFAR100~\cite{krizhevsky2009learning} dataset. 
A neural classifier is pre-trained on 50 randomly selected image classes.
Then during the \textit{incremental learning phase}, the classifier is trained on an \textit{incremental training set} containing all 100 classes (with 50 new classes not in the initial training phase) to evaluate how quickly it acquires new knowledge. 

             \begin{table*}
    \setlength{\tabcolsep}{7pt}
            \centering
            \begin{tabular}{lccccccccc}
                  \toprule
                  & \multicolumn{3}{c}{ResnetV1} & \multicolumn{3}{c}{Densenet}  & \multicolumn{3}{c}{MobilenetV2} \\
                  \cmidrule(r){2-4} \cmidrule(r){5-7} \cmidrule(r){8-10} 
                 Model  & New & Old & Overall & New & Old & Overall & New & Old & Overall \\
                 \cmidrule(r){1-1} \cmidrule(r){2-4} \cmidrule(r){5-7} \cmidrule(r){8-10} 
                  \Parametric & 35.82\%  & 73.16\% & 54.49\% & 33.18\%  & 72.58\% & 52.88\%  & 34.12\%  & 70.24\% & 52.18\%  \\
                  \Mixture & 36.34\%  & \textbf{73.68}\% & 54.91\% & 33.04\%  & \textbf{72.94}\% & 52.99\%  & 34.64\%  &\textbf{70.38}\% & 52.42\% \\
                  \MbPA & 38.56\%  & 73.56\% & 56.06\% & 34.42\%  & 72.39\% & 53.33\%  & 35.54\%  & 69.74\% & 52.64\%  \\
                  \Hebb & \textbf{41.46}\%  & 73.16\% & \textbf{57.31}\% & \textbf{37.08}\%  & 72.18\% & \textbf{54.38}\%  & \textbf{38.16}\%  & 69.32\% & \textbf{53.74}\% \\
                  \bottomrule
            \end{tabular}
            \caption{
                Average Top-1 accuracy for the online image classification experiment. For each base model, Top-1 accuracy on 50 new classes (\textbf{New}) in testing, 50 old classes (\textbf{Old}) in pre-training, and all 100 classes (\textbf{Overall}) are reported. 
            } \vspace{-0.02in}
            \label{table:cifar-online}
      \end{table*}

                \begin{figure*}[htb!]
            \centering       
            \includegraphics[width=0.315\textwidth]{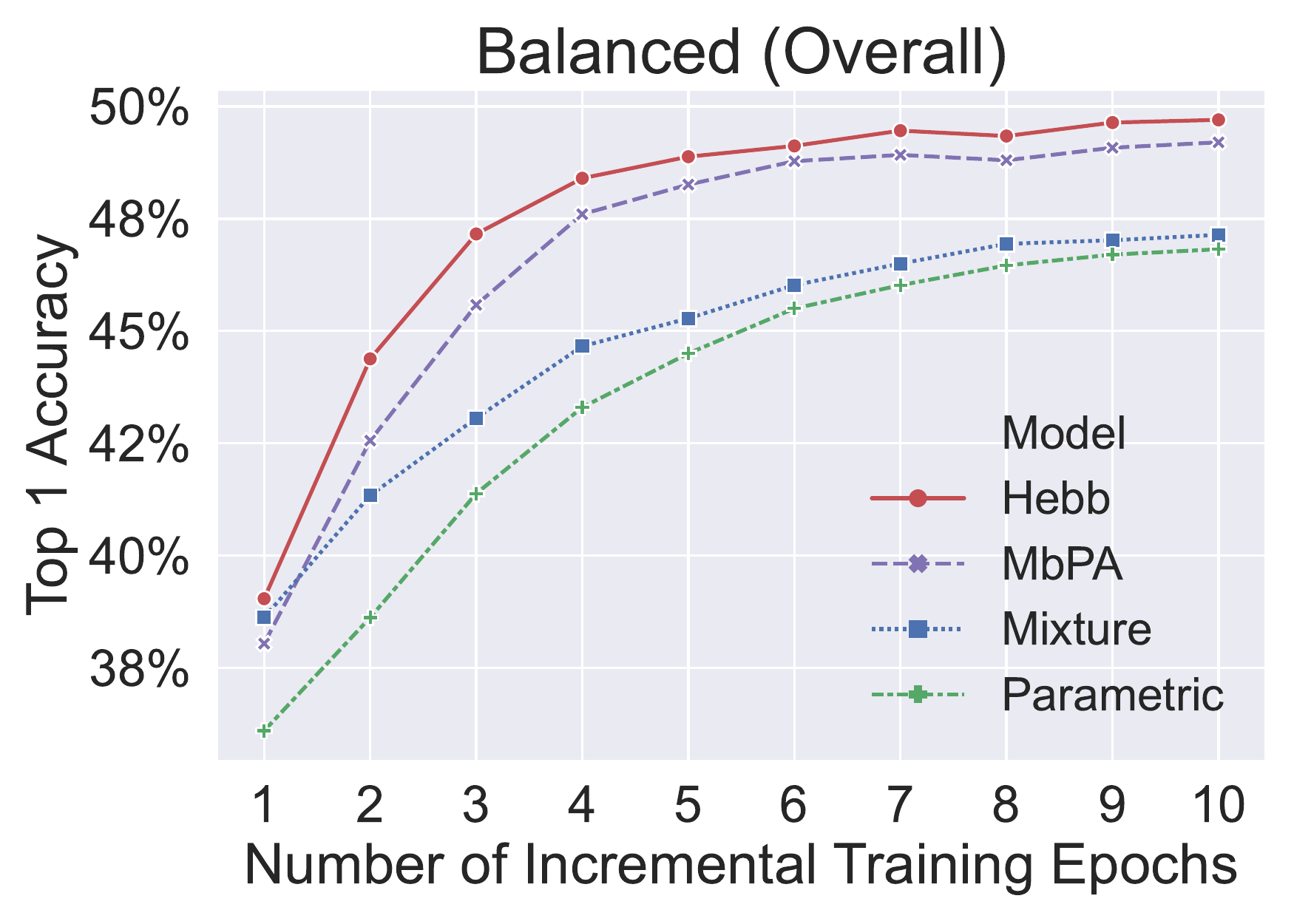}  \quad
            \includegraphics[width=0.315\textwidth]{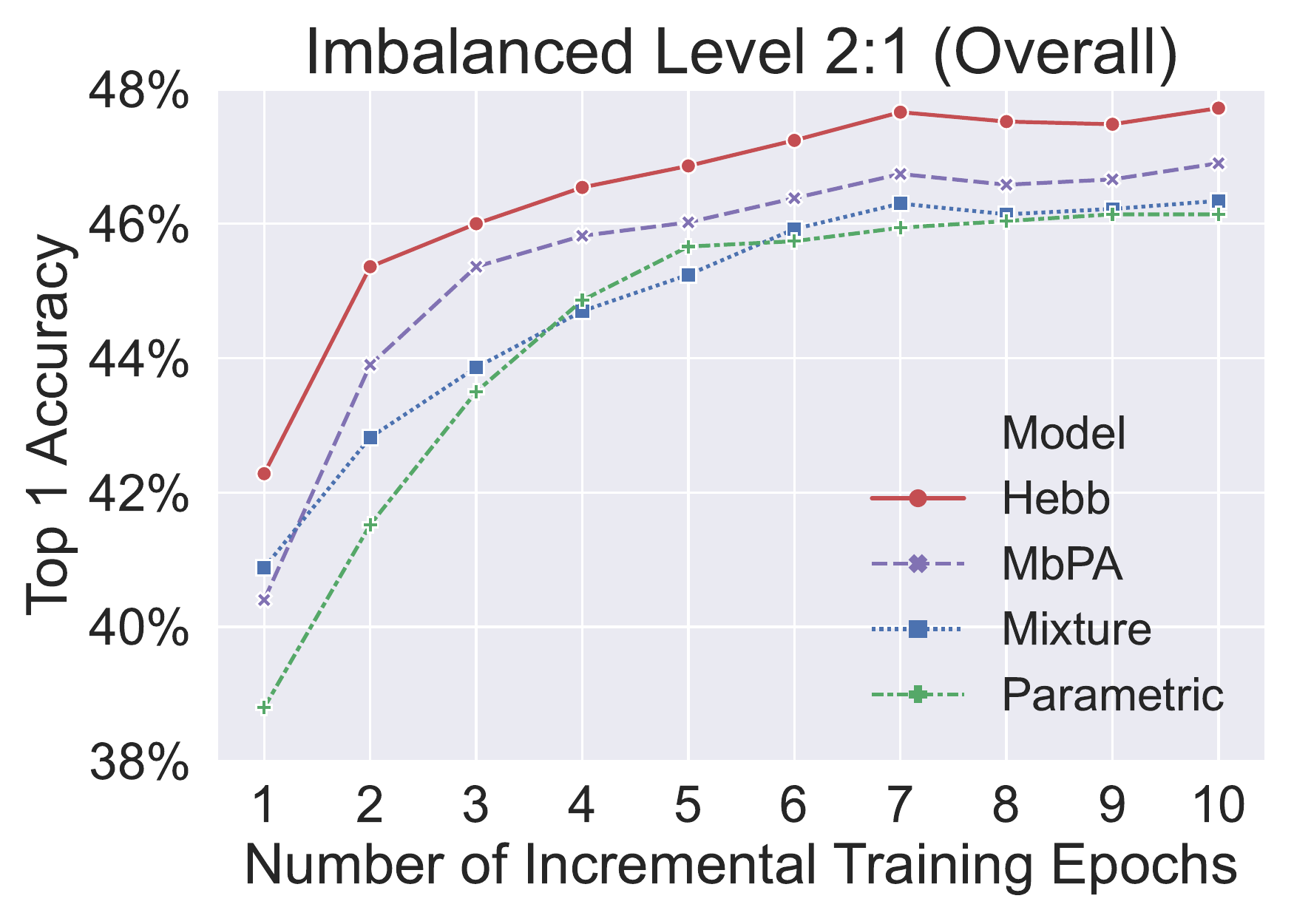} \quad
            \includegraphics[width=0.315\textwidth]{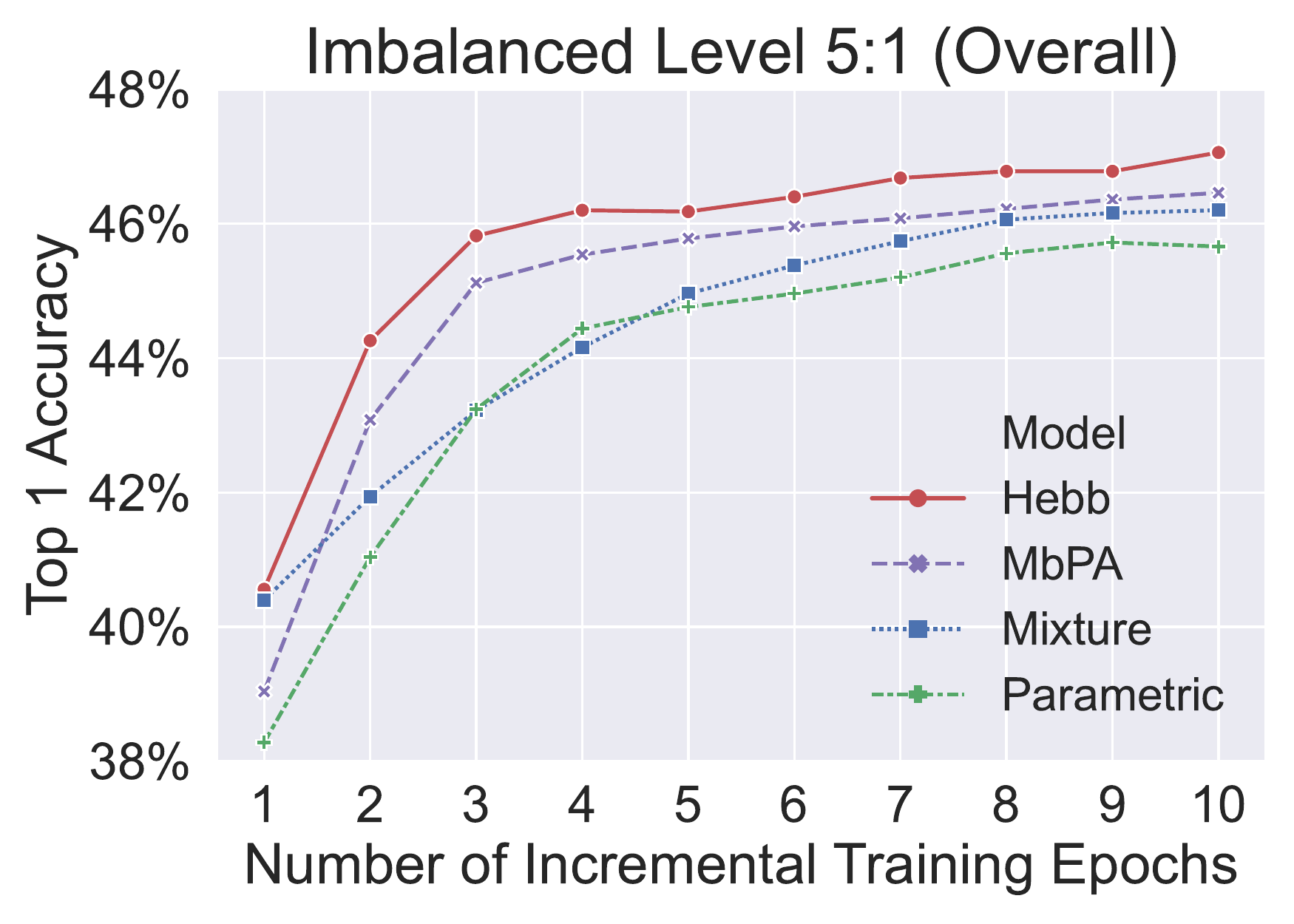}          
            
           \caption{
                Top-1 accuracy of different methods at different incremental training epochs in the incremental image classification scenario. The balanced incremental learning setting (\textbf{Left}) and two imbalanced incremental learning settings (\textbf{Middle} and \textbf{Right}) with two different imbalance levels are included.
                } 
            \label{fig:cifar-incre}
      \end{figure*}
      
 We tested two big networks: ResnetV1 \cite{he2016deep} with 50 layers and Densenet~\cite{huang2017densely}, and a lightweight network MobilenetV2~\cite{sandler2018mobilenetv2}.
     The pre-training phase is the same, and several baselines are compared during the incremental learning phase:
      
      \textbf{\Parametric}: It fine-tunes the model on the incremental training set without using the memory component.

      \textbf{\Mixture} \cite{grave2016improving,grave2017unbounded,tu2018learning,orhan2018simple}: It combines the prediction of \Parametric\ with a non-parametric prediction from neighbors $\mathbf{N}$ by:
      \begin{equation}
          P(y) \propto (1 - \gamma)   e^{ f_{\omega_y} (\mathbf{h}_t) } +  {\gamma \sum_{k=1}^{|\mathbf{N}|} \mathds{1}(y_k = y) e^{\theta \mathbf{h}_k^{T} \mathbf{h}_t}} ,
      \end{equation}
      where $\mathds{1}(y_k = y) $ is the an indicator function, $\gamma$ controls the contribution of each part, and $\theta$ controls the flatness of the non-parametric prediction.
      
      \textbf{\MbPA} \cite{sprechmann2018memory}: It adapts the output network of \Parametric~using gradient descent to maximize Eq. (\ref{eq:mbpa1}) w.r.t. neighbors $\mathbf{N}$. 
    
    \textbf{\Hebb~(proposed)}: Our scheme (c.f. Algorithm 1) dynamically combines the proposed Hebbian update with \MbPA~to adapt the output network of \Parametric. 
    
      \paragraph{Class Balanced Incremental Learning}
      Test accuracy of different methods on all 100 classes and 50 new classes are reported in Table \ref{table:cifar-incre}. Performances on 50 old classes are not presented due to limited variations among different methods.
      Two interesting observations can be noted:
      \begin{itemize} 
      \item \Hebb~achieves the best overall accuracy on all 100 classes at all epochs. 
      Although \MbPA~notably outperforms both \Parametric~and \Mixture, \Hebb~consistently outperforms \MbPA~at all epochs.
\item \Hebb~learns new classes \textit{better}. 
\Hebb~has evident improvements on 50 new classes, indicated by 3-5\% gain over \MbPA~and 8-10\% gain over \Mixture~and \Parametric.
\item \Hebb~learns new classes \textit{faster}. 
The optimal accuracy on new classes achieved by \Parametric~and \Mixture~at epoch 10 can be obtained by \Hebb~within 3 epochs.
      \end{itemize}

        \paragraph{Class Imbalanced Incremental Learning}
    
    CIFAR100 and most other datasets are artificially balanced; however, data imbalance is inevitable in most real-world applications.
      In this experiment, we follow the incremental learning experiment setup in Section \ref{subsec:incre} using ResnetV1, and we constructed two imbalanced incremental training sets.
      \textit{Imbalanced level 2:1}: half of 50 new classes have twice samples as many as the other half. \textit{Imbalanced level 5:1}: half of 50 new classes have five times samples as many as the other half.
      Models are still evaluated on the balanced test set of all 100 classes.
      The performances of different methods using ResnetV1 are presented in Figure~\ref{fig:cifar-incre} (Middle and Right). 
      \MbPA~outperforms \Parametric~and \Mixture~with notable margins in the balanced setup (Figure~\ref{fig:cifar-incre}-Left). However, the improvement margin is degraded in these two imbalanced setups.
     In contrast, \Hebb~is still \emph{consistently and notably better} than \MbPA\ at all epochs.
     This result reveals that \Hebb~is well suited to learn imbalanced new classes.

      \subsection{Online Adaptation for Image Classification}
\label{subsec:cifar_online}
      This online setting aims to evaluate the ability of a pre-trained model to learn new classes in an online manner.
     The pre-training phase is the same as the previous incremental learning experiment, in which the three base neural classifiers trained on 50 randomly sampled classes of CIFAR100. 
     During online testing, we sequentially feed the complete test set with all 100 classes.
     The base classifier (\Parametric) and the memory module used by \Mixture, \MbPA, and \Hebb\ are updated as every 100 test samples arrive. 
     
    The average Top-1 accuracy after the online testing phase is summarized in Table~\ref{table:cifar-online}. Different methods perform similarly on 50 old classes, with the simple \Mixture~method being slightly better.
    \Hebb~achieves the best overall performance on all 100 classes. It outperforms the closest (and SOTA) competitor \MbPA\ by more than 1\% and other baselines by larger margins.
Furthermore, \Hebb~is especially good at learning 50 new classes.
      It outperforms \MbPA~by 2-3\% and outperforms \Mixture~ and \Parametric~by 3-5\%.

     \begin{table}[t]
        \centering
        \begin{tabular}{lccc}
              \toprule
              & Pre-train 30 & Pre-train 50  & Pre-train 70 \\
              \cmidrule(r){2-2} \cmidrule(r){3-3} \cmidrule(r){4-4} 
              \Parametric  & 40.97\% & 54.49\%  & 66.42\%  \\
              \Mixture & 41.47\% & 54.91\% &  66.74\% \\
              \MbPA & 43.50\% & 56.06\% & 67.39\%  \\
              \Hebb & \textbf{45.66}\% &  \textbf{57.31}\% & \textbf{68.37}\%\\
              \bottomrule
        \end{tabular}
        \caption{Top-1 accuracy on all 100 classes of using different number of pre-training classes (30, 50, 70) in the online adaptation experiment for image classification.} 
        \label{table:pretrain}
  \end{table} 
  
     \paragraph{Varying Number of Pre-training Classes}

  In this experiment, we follow the online adaptation setup in Section \ref{subsec:cifar_online}, yet we vary the number of pre-training classes.
Apart from 50 pre-training classes, we additionally report in Table \ref{table:pretrain} the overall performances of using 30 and 70 pre-training classes trained with ResnetV1.
The fewer classes used for pre-training, the more new classes need to be captured during the online testing phase.
We can see from  Table \ref{table:pretrain} that \Hebb\ is
consistently the best with a different number of pre-training classes.
Furthermore, its improvement margin increases as more number of new classes need to be captured (e.g., when the number of pre-training classes decreases from 70 to 30). This result reinforces our conclusion that \Hebb\ is especially good at learning new class patterns quickly.

For online image classification, we also include three extra experiments in Appendix B: (1) the hyper-parameter sensitivity of \Hebb; (2) the computation efficiency of \Hebb; (3) an ablation study analyzing the effect of different components of \Hebb.

      \subsection{Online Adaptation for Language Model}
\label{subsec:lm}
      
      Finally, we studied an online adaptation setting for the language model task to capture new vocabularies or distributions during test time.
      Two benchmark datasets are used, i.e., Penn Treebank (PTB)~\cite{marcus1993building} and WikiText-2~\cite{merity2016pointer}. 
      PTB is relatively small with vocabulary size 10,000.  WikiText-2 from Wikipedia articles is larger with vocabulary size 33,278. 
  
      We consider two types of testing data.    
      In the first \textbf{intra-domain} scenario, the testing data come from the same domain as the training data for pre-training with slight word distribution shifts. Because no new vocabularies are encountered, the Hebbian update of \Hebb~is computed over all entries in $\mathbf{N}$.
      In the second \textbf{cross-domain} scenario, models are pre-trained on the training data of WikiText-2 and evaluated on the test set of PTB. This scenario contains both domain shifts and 3.77\% out-of-vocabulary (OOV).
    We use a state-of-the-art \emph{LSTM} (AWD-LSTM)~\cite{merity2017regularizing} as the base neural model.
    It is fixed during testing in the intra-domain scenario, and it is updated continually for every mini-batch (100 tokens in our case) in the more challenging cross-domain scenario. 
      We reported perplexity (ppl.) and cross-entropy loss (CE-loss) for the two scenarios, respectively, because perplexities ($\text{exp}^{\text{CE-loss}}$) on OOV in the cross-domain scenario are too large to be compared.

            \begin{table}[t!]
      \small
          \setlength{\tabcolsep}{3.7pt}
            \centering
            \begin{tabular}{lcccc|cc}
                  \toprule
                  & \multicolumn{4}{c}{Intra-domain} & \multicolumn{2}{|c}{Cross-domain} \\
                  \cmidrule(r){2-5} \cmidrule(r){6-7} 
                  & \multicolumn{2}{c}{PTB} & \multicolumn{2}{c}{WikiText-2} & \multicolumn{2}{|c}{PTB} \\
                  \cmidrule(r){2-3} \cmidrule(r){4-5} \cmidrule(r){6-7}
                  & Valid & Test &  Valid & Test &  Overall & OOV  \\
                  \texttt{LSTM}  & 60.88 & 58.53 & 68.50 & 65.44 &6.41 & 12.66 \\
                  \Mixture & 58.37 & 57.44 & 54.80  & 52.50 &  5.72 & 9.74 \\
                  \MbPA    & 59.00 & 56.68& 61.98  & 59.00 & 6.12 & 12.04 \\
                  \Hebb & 58.45 & 56.39 & 60.95 & 57.29 & 5.99 & 10.36 \\
                  \Mixture+\MbPA & 56.95 & 55.58 & 53.00 & 50.21  & 5.68 & 9.70 \\
                  \Mixture+\Hebb & \textbf{56.30} & \textbf{55.13} & \textbf{52.65} & \textbf{49.66} &  \textbf{5.61} & \textbf{9.65} \\
                  \bottomrule
            \end{tabular} 
            \caption{
            Results of online adaptation experiments for the language model task. \textbf{Left}: intra-domain results on PTB and Wikitext-2 evaluated by perplexity. \textbf{Right}: cross-domain results on PTB with \texttt{LSTM} pretrained on Wikitext-2 evaluated by cross-entropy loss. \vspace{-0.05in}
            }
            
            \label{table:lm}
      \end{table}

    \begin{table}[t]
            \centering
            \small
            \begin{tabular}{lcccccc}
                  \toprule
                  & $\!<\!$ 50 & 50-100 & 100-500 & $\!>\!$ 500 & All\\
                  \cmidrule(r){2-6} 
            \textbf{PTB} & & & & \\
                  \MbPA & 3061.75  & 719.73 & 189.96 & 13.64 & 56.68 \\
                  \Hebb & 2840.84  & 655.86 & 177.61 & 13.13 & 56.39\\
            \textbf{WikiText-2} & & & & \\
                  \MbPA & 5027.67  & 970.07 & 277.40 & 12.53 & 59.00\\
                  \Hebb & 4676.85  & 914.83 & 264.13 & 12.36  & 58.29\\
                  \bottomrule
            \end{tabular}
            \caption{Test perplexity versus word appearing frequency in the intra-domain scenario.} 
            \label{table:frequency}
      \end{table}

      Table~\ref{table:lm} summarizes the results of these two scenarios, and Table \ref{table:frequency} further presents the test perplexity broken down by word frequency in the intra-domain scenario.
      Two observations can be noted.
\textbf{First}, \Hebb~consistently achieves better overall performance than \MbPA. 
The overall performance gain of \Hebb~over \MbPA~is mainly obtained from OOV in the cross-domain scenario and the less frequent vocabularies in the intra-domain scenario by inspecting Table \ref{table:frequency}. 
This observation validates that \Hebb~is especially effective to learn new vocabularies (OOV) or infrequent vocabularies.
      \textbf{Second}, \emph{Mixture+Hebb} achieves the best performance. Although the simple \Mixture~method is very strong and it outperforms both \MbPA~and \Hebb\ in most cases, hybridizing it with \MbPA~(\emph{Mixture+MbPA}) or with \Hebb~(\emph{Mixture+Hebb}) can consistently boost its performance. It means that \MbPA~and \Hebb~both have \textit{orthogonal benefits} when combined with \Mixture.

      \section{Conclusion}

 This paper considers scenarios that require learning new classes or data distributions quickly without forgetting previous ones. 
    To tackle the two major challenges (catastrophic forgetting, sample efficiency) towards this goal, we propose a method called ``Memory-based Hebbian Parameter Adaptation'' (\Hebb).
    \Hebb\ augments a regular neural classifier with a continuously updated memory module and a new parameter adaptation method based on the well-known Hebbian theory.
    Extensive experiments on a wide range of learning tasks (image classification, language model) and learning scenarios (continual, incremental, online) demonstrate the superior performance of \Hebb.

\bibliographystyle{named}
\bibliography{aaai20}

 \clearpage

\appendix

\part*{Technical Appendix}

  \begin{figure*}[!ht]
		\centering
		\includegraphics[width=0.99\textwidth]{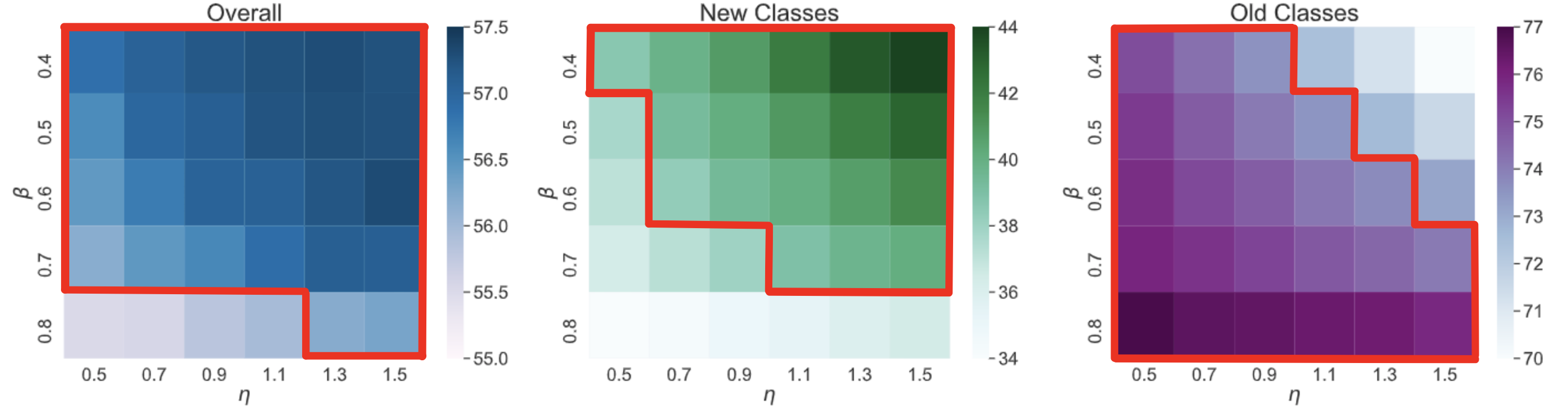} 
 \caption{ Hyperparameter sensitivity analysis of Hebb for the online image classification experiment on CIFAR100 using ResnetV1.  \textbf{(Left)}: overall Top 1 accuracy on 100 classes; \textbf{(Middle)}: Top 1 accuracy on 50 new classes; \textbf{(Right)}: Top 1 accuracy on 50 old classes. Cells within the red polygon are better than \MbPA.}  
        \label{fig:hyper}
\end{figure*}

\section{Reproducibility Checklist}
\label{appendix:repo_checklist}

\subsection{Model Details and Hyper-parameters for Image Classification Experiments}

\subsubsection{Incremental Learning \& Online Adaptation}
During the initial model pre-training phase, different base neural classifiers (ResnetV1, MobilenetV2, and Densenet) are trained with SGD with momentum 0.9, learning rate 5e-4, and batch size 128. In total 350 epochs are trained, with a learning rate 0.1 in the first 150 epochs, 0.01 in the next 100 epochs, and 0.001 in the last 100 epochs.

After pre-training the neural classifiers, hyper-parameters of different baseline methods are tuned for the two learning scenarios (online and incremental) to maximize the overall performance on 100 classes. The hyper-parameter search space for different methods are:

\begin{itemize}
\item \Parametric: We use RMSprop optimizer and tune the learning rate Lr $\in \{5e^{-5}, 1e^{-4}, 5e^{-4}, 1e^{-3}\}$.
\item \Mixture: Two weights $\theta \in \{0.4, 0.6, 0.8, 1 \}$ and $\gamma \in \{0.05, 0.1, 0.2, 0.3 \}$ are tuned and.
\item \MbPA:  The learning rate $\lambda  \in \{0.01, 0.02, 0.05, 0.1, 0.2\}$ and the number of optimization steps $\in \{1, 5, 10\}$ of the RMSprop optimizer are tuned without weight decay.
\item \Hebb: The learning rate $\eta  \in \{0.5, 0.7, 0.9, 1.1, 1.3, 1.5\}$ and $\beta \in \{0.4, 0.5, 0.6, 0.7, 0.8 \}$ of the dynamic weight term $E_y$ are tuned. It also re-uses the optimal hyper-parameters of \MbPA.
\end{itemize}
The optimal hyper-parameters of different models and settings are presented in Table \ref{table:cifar_parameter}.

\subsubsection{Continual Learning}
We use the Adam \cite{kingma2014adam} as the optimizer with learning rate 1e-3 for MLP. The regularization term of EWC is set to be 1000. For \MbPA, $\lambda$ is set to be 0.05, and the optimization step is set to 5. For \Hebb, $\eta$  is set to 0.2 and $\beta$ is set to 0.9.

\subsection{Model Details and Hyper-parameters for Language Model Experiments}

For the base AWD-LSTM, we used 3 LSTM layers with a size 1200 each.
For the initial model pre-training on PTB, the batch size is set to 20, the input layer dropout is set to 0.4, the hidden layer dropout is set to 0.25, and 500 epochs are trained. 
For the initial model pre-training on Wikitext-2, hidden layer dropout is set to 0.2, and 750 epochs are trained. 
Other configurations not mentioned are set by default according to the official implementation \footnote{\url{https://github.com/salesforce/awd-lstm-lm}}.

After the initial model pre-training phase, different methods are evaluated to learn distribution shifts or OOVs continually. We tune the hyper-parameters of different methods on corresponding validation sets to maximize overall perplexity on all vocabularies for both intra-domain and cross-domain scenarios.
The hyper-parameter spaces to search for \emph{LSTM}, \Mixture, \MbPA~and \Hebb~are slightly different from previous image classification experiments.
For the two additional methods that hybridize \Mixture~with \MbPA
~(\emph{Mixture+MbPA}) and with \Hebb~(\emph{Mixture+Hebb}), we tune the weight ($w$) multiplied to \Mixture~and the other $1-w$ fraction is multiplied to \MbPA~or to \Hebb. 
The optimal hyper-parameters of different methods and setting are summarized in Table \ref{table:lm_parameter}.

\begin{table}
\centering
\setlength{\tabcolsep}{0.6pt}
\small
\begin{tabular}{c cccccccc}
\toprule
 &  & \Parametric & \multicolumn{2}{c}{\Mixture}                   & \multicolumn{2}{c}{\MbPA}      & \multicolumn{2}{c}{\Hebb}                   \\
  \cmidrule(r){3-3} \cmidrule(r){4-5} \cmidrule(r){6-7} \cmidrule(r){8-9} 
&  &      Lr         & $\theta$ & $\gamma$ & $\lambda$ & step & $\beta$ & $\eta$ \\
\midrule

\multirow{3}{*}{\makecell{\textbf{Incremental} \\ \textbf{Learning}}}  &  ResnetV1    & 5e-4     & 1                     & 0.1                   &1e-3     & 5    & 0.6                 & 1.5                 \\
&   Densenet    &     1e-3       &           1            &           0.05            &      1e-3  &  10     &         0.5             &            0.5         \\
&   MobilenetV2 & 5e-4    & 1                     & 0.05                  & 1e-3  & 5    & 0.5                  & 0.5                 \\
\midrule

\multirow{3}{*}{\makecell{\textbf{Online} \\ \textbf{Adaptation}}} &   ResnetV1    & 1e-4     & 1.0                     & 0.1      & 1e-4     & 5    & 0.5    & 1.5                   \\
&   Densenet    &     5e-4       &    0.8     &           0.1     &        1e-4         &  5     &         0.6         &       1.1         \\
&   MobilenetV2 & 1e-4    & 0.8     & 0.05      &  1e-4   & 5    & 0.5   & 1.1          \\

\bottomrule
\end{tabular}
\caption{Optimal hyper-parameters for different methods for image classification experiments in both incremental learning and online adaptation scenarios.}
\label{table:cifar_parameter}
\end{table}

\begin{table}
\setlength{\tabcolsep}{1.8pt}
\small
\centering
\begin{tabular}{lccc}
\toprule
& \multicolumn{2}{c}{Intra-domain} & Cross-domain\\
\cmidrule(r){2-3} \cmidrule(r){4-4} 
& \textbf{PTB} & \textbf{WikiText-2} & \\
\emph{LSTM} (Lr) & - & - & 1e-4  \\
\Mixture~($\theta, \gamma$) & $(0.4,  0.01)$ & $(0.4, 0.15) $ & (0.4,0.3)  \\
\MbPA~($\lambda, step$) & $(0.1, 5)$ & $(0.1, 5)$& (2, 1)  \\
\Hebb~($\beta, \eta$)  & $(0.6, 0.7)$ & $(0.15,  0.4)$ & (0.5,0.3) \\
\emph{Mixture+MbPA} ($w$) & 0.05 & 0.15 & 0.3 \\
\emph{Mixture+Hebb} ($w$) & 0.05 & 0.2 & 0.3 \\
\bottomrule
\end{tabular}
\caption{Optimal hyper-parameters for different methods for language modelling experiments in both intra-domain and cross-domain settings.}
\label{table:lm_parameter}
\end{table}

\section{Additional Experiment Results}

\subsection{Hyper-parameter Sensitivity of \Hebb}
\label{appendix:hyper_sensitivity}
We present a hyper-parameter sensitivity analysis of \Hebb in the online image classification experiment.
We can see from Figure \ref{fig:hyper} (Left) that the overall performance of Hebb is not sensitive to these two hyper-parameters: only four configurations out of the red polygon are slightly worse than \MbPA. 
These two hyper-parameters mainly affect the performance distributed to new and old classes, as illustrated in Figure \ref{fig:hyper} (Middle and Right). As $\eta$ increases and $\beta$ decreases, the performance on new classes increases, while the performance on old classes drops. 
Results reported in this paper were tuned to maximize the overall performance. However, readers can easily set these two directional hyper-parameters to favor either new or old classes.

\subsection{Computation Efficiency of \Hebb}

\label{appendix:computation}

\begin{table}[t!]
        \centering
        \begin{tabular}{lccc}
              \toprule
              & \multicolumn{1}{c}{ResnetV1} & \multicolumn{1}{c}{Densenet}  & \multicolumn{1}{c}{MobilenetV2} \\
              \cmidrule(r){2-2} \cmidrule(r){3-3} \cmidrule(r){4-4} 
              \Parametric  & 45.5 & 40.3  & 35.5   \\
              \Mixture & 51.4 & 44.5  & 40.7  \\
              \MbPA & 65.5 & 56.8  & 52.5 \\
              \Hebb & 67.7 & 59.5  & 54.7  \\
              \bottomrule
        \end{tabular}
        \caption{ The computation time (in seconds) of one online testing run for the online image classification experiment.}
        \label{table:time}
  \end{table}

In this experiment, we include the computation time of different methods of the online image classification experiment on CIFAR100 in Table \ref{table:time}. All methods are computed using a single GPU (GeForce GTX TITAN X). We can see that the computation overhead of Hebb on top of \MbPA\ is marginal (within 5\% increase).
The computation efficiency of \Hebb\ can be explained threefold:
\begin{enumerate}
\item No additional forward pass is needed; free feature representations computed by the forward pass of \MbPA. 
\item No additional retrieval of nearest neighbors is needed; \ MbPA already retrieves nearest neighbors.
\item No extra backward pass is needed by \Hebb, and only a few additions and multiplications in Eq.~\eqref{eq:hebbrule} and Eq.~\eqref{eq:interpolation} are required. 
\end{enumerate}

\subsection{Ablation study of \Hebb}
\label{appendix:ablation}
      
In this experiment, several simplified versions of \Hebb~are tested and compared to justify our design choices. 
\begin{itemize}
\item \emph{Hebb-v1}: It only uses the Hebbian update in Eq.~\eqref{eq:hebbrule} to adapt the output network's parameters w.r.t. $\mathbf{N^{new}}$; the gradient-based \MbPA~update is discarded. 
\item \emph{Hebb-v2}: It differs from \Hebb~by computing the Hebbian update w.r.t. $\mathbf{N}$, rather $\mathbf{N^{new}}$.
\item \emph{Hebb-v3}: This version differs from \Hebb~by using a fixed static weight, rather than the dynamic weighting term $E_y$, to merge $\Delta^{Hebb}_{\omega_y}  $ and $\Delta^{MbPA}_{\omega_y}$. 
\item \emph{Hebb-250/500}: This version implements a fixed-size memory by a \textit{ring buffer} to store a
limited number (250/500) of latest testing data. 
\end{itemize}

            \begin{table}[t]
                \setlength{\tabcolsep}{5pt}
            \centering
            \begin{tabular}{lcccccc}
                  \toprule
                  & New & Old & Overall \\
                  \cmidrule(r){2-2} \cmidrule(r){3-3} \cmidrule(r){4-4} 
                  \MbPA & 38.56\%  & \textbf{73.56}\% & 56.06\% \\
                  \emph{Hebb-v1} & 39.58\%  & 71.04\% & 55.31\%  \\
                  \emph{Hebb-v2} & 40.02\%  & 72.64\% & 56.33\% \\
                  \emph{Hebb-v3} & 40.66\%  & 72.84\% & 56.75\% \\
                  \emph{Hebb-250} & 39.91\%  & 72.31\% & 56.11\% \\
                  \emph{Hebb-500} & 40.20\%  & 72.45\% & 56.40\% \\
                  \Hebb & \textbf{41.46}\%  & 73.16\% & \textbf{57.31}\% \\
                  \bottomrule
            \end{tabular}
            \caption{Ablation study comparing \Hebb~with simplified versions in the online adaptation experiment using ResnetV1. Results on 50 new classes, 50 old classes, and all 100 classes are reported separately}  
            \label{table:ablation}
      \end{table}
      
      Several observations can be noted from Table~\ref{table:ablation}.
      \textbf{First}, the three variants (i.e. \emph{Hebb-v2}, \emph{Hebb-v3}, \Hebb) that combines MbPA update (\MbPA) with Hebbian update (\emph{Hebb-v1}) all outperform the individual update variants. 
  This result shows that combining these two update rules is better than using them separately. 
      \textbf{Second}, the superior performance of \Hebb~over \emph{Hebb-v2} reveals the benefit of computing the Hebbian update w.r.t. only $\mathbf{N^{new}}$.
      \textbf{Third}, the superior performance of \Hebb~over \emph{Hebb-v3} justifies the effectiveness of the proposed dynamic interpolation in eq.~\eqref{eq:interpolation}.
      \textbf{Fourth}, the superior performance of \Hebb~over \emph{Hebb-250/500} shows the benefit of using large memory sizes. Nevertheless, \emph{Hebb-200/500} (with limited memory size) is sufficient to outperform \MbPA\ (unlimited memory size), especially on new classes.

\end{document}